%% file: main.tex
\documentclass[runningheads]{llncs}
\usepackage[T1]{fontenc}
\usepackage{graphicx}
\usepackage[linesnumbered,ruled,vlined]{algorithm2e} 
\usepackage{subcaption}
\usepackage{import}
\usepackage{multirow}
\usepackage{graphicx}
\usepackage{xcolor}
\usepackage{epsfig,endnotes}
\usepackage{import}
\usepackage{url}            
\usepackage{commath}
\usepackage{booktabs}       
\usepackage{amsfonts}       
\usepackage{nicefrac}       
\usepackage{microtype}      
\usepackage{array}
\usepackage{stfloats} 
\usepackage{epigraph}
\usepackage{tikz}
\usepackage{amsmath}

\graphicspath{{./images/}}

\usepackage{color,soul}

\usepackage{arydshln}
\usepackage{enumitem}
\setlist{itemsep=1pt}
\setlist[itemize]{leftmargin=5mm}
\setlist[enumerate]{leftmargin=5mm}
\usepackage[font=small,skip=3pt]{caption}
\usepackage{titlesec}
\titlespacing*{\subsection}{0pt}{2.5ex plus 1ex minus .2ex}{3pt}
\titlespacing*{\subsubsection}{0pt}{1.5ex plus 1ex minus .2ex}{2pt}
\usepackage{float}
\usepackage{adjustbox}
\newcommand\blfootnote[1]{%
  \begingroup
  \renewcommand\thefootnote{}\footnote{#1}%
  \addtocounter{footnote}{-1}%
  \endgroup
}
\usepackage{macros}

\begin{document}
%
\title{\name}
\titlerunning{PassGPT: Password modeling and generation with LLMs}
%
\author{
    Javier Rando\inst{1}\orcidID{0000-0002-2723-7660} \and
    Fernando Perez-Cruz\inst{1,2}\orcidID{0000-0001-8996-5076} \and
    Briland Hitaj\inst{3}\orcidID{0000-0001-5925-3027}
}

%
\authorrunning{J. Rando et al.}
%
\institute{
    ETH Z\"urich, Turnerstrasse 1, 8092 Zürich\\
    \email{jrando@ethz.ch} \and
    Swiss Data Science Center, Turnerstrasse 1, 8092 Zürich
    \email{fernando.perezcruz@sdsc.ethz.ch} \and
    SRI International, New York, NY 10165 USA\\
    \email{briland.hitaj@sri.com}
}
\maketitle              

\import{\sectiondir}{01-abstract.tex}
%
%
%
\import{\sectiondir}{02-introduction.tex}
\import{\sectiondir}{03-background.tex}
\import{\sectiondir}{04-experimental-setup.tex}

\import{\sectiondir}{05-evaluation.tex}

\import{\sectiondir}{07-conclusions.tex}


%
%
%
\bibliographystyle{splncs04}
\bibliography{main}

\appendix
\import{\sectiondir}{10-appendix.tex}

\end{document}

%% file: sections/01-abstract.tex
\begin{abstract}
Large language models (LLMs) successfully model natural language from vast amounts of text without the need for explicit supervision. In this paper, we investigate the efficacy of LLMs in modeling passwords. We present \textbf{PassGPT}, an LLM trained on password leaks for password generation. PassGPT outperforms existing methods based on generative adversarial networks (GAN) by guessing twice as many \emph{previously unseen} passwords. Furthermore, we introduce the concept of \emph{guided password generation}, where we leverage PassGPT sampling procedure to generate passwords matching arbitrary constraints, a feat lacking in current GAN-based strategies. Lastly, we conduct an in-depth analysis of the entropy and probability distribution that PassGPT defines over passwords and discuss their use in enhancing existing password strength estimators.\blfootnote{
Code and models can be accessed at \url{https://github.com/javirandor/passgpt}}

\keywords{Password Guessing \and LLMs \and Generative AI}
\end{abstract}

%% file: sections/02-introduction.tex
\section{Introduction}
\label{sec:introduction}

Passwords still retain their status as the authentication mechanism of choice despite the ever-increasing number of alternative technologies~\cite{wayman2005biometric,paterson2010one}, primarily thanks to passwords being easy to deploy and remember. Furthermore, most applications rely on passwords as a fallback mechanism when other methods do not succeed.
Considering their prevalence, password leaks~\cite{wiredLeak2019,techrepublic2021} are one of the main threats institutions (and individuals) face. Not only do password leaks enable adversaries to break into systems but they also make it possible to research and identify hidden patterns within human-generated passwords that guide creation and refinement of effective password cracking tools~\cite{hashcatweb,johntheripperweb}.

Machine learning (ML) has played (and continues to play) a prominent role in extracting and learning meaningful features from vast password leaks, resulting in major contributions primarily towards two main areas of research: (1) password guessing~\cite{melicher2016fast,hitaj2019passgan,pagnotta2022passflow,pasquini2021reducing,pasquini2020interpretable} and (2) password strength estimation mechanisms~\cite{ur2012does,melicher2016fast,de2014very,golla2016security,golla2018accuracy}.

At the same time, Large Language Models (LLMs), a family of ML models, has demonstrated tremendous effectiveness in natural language processing (NLP) and understanding (NLU). These models are based on the \emph{Transformer} architecture; well-known examples include the Generative Pre-trained Transformer (GPT) models~\cite{brown2020language,chatgpt}, PaLM \cite{chowdhery2022palm} or LLaMA \cite{touvron2023llama}.
Given their recent success, we pose the following question: \textbf{How effectively can \emph{LLMs} capture the underlying characteristics and cues hidden in the complex nature of human-generated passwords?}

To answer this question, we present and thoroughly evaluate an LLM-based password-guessing model called \textbf{PassGPT}. Suitable for both password guessing and password strength estimation, PassGPT is an \emph{offline} password-guessing model based on the GPT-2 architecture~\cite{radford2018improving}. When compared with prior work on deep generative models~\cite{hitaj2019passgan,pasquini2020interpretable}, PassGPT guesses \textbf{$20\%$} more \emph{unseen passwords}, and demonstrates good generalization capabilities to novel leaks. Moreover, we enhance PassGPT with vector quantization~\cite{yu2021vector}. The resulting architecture is \textbf{PassVQT}, which can increase the perplexity of generated passwords.

Unlike previous deep generative models that generate passwords as a whole, PassGPT sequentially samples each character,thus introducing the novel task of \emph{guided password generation}. This method ensures a more granular (character-level) guided exploration of the search space, where generated passwords are sampled based on arbitrary constraints.
Finally, PassGPT, in constrast with GANs, provides an explicit representation of the probability distribution over passwords. We show that password probabilities align with state-of-the-art password strength estimators: PassGPT assigns lower probabilities to stronger passwords. We also look for passwords that can be easily guessed by generative approaches, even though they are considered "strong" by strength estimators. 
We discuss how password probabilities under PassGPT can be valuable for enhancing existing strength estimators.

\subsection{Contributions and Remarks}

Given the nature of LLMs, as well as the probabilistic nature of DL-based password generation, we position ourselves in \textbf{offline password guessing}. Such a setup is in line with prior work in the domain~\cite{hitaj2019passgan,pasquini2021improving,pasquini2021reducing,pagnotta2022passflow,durmuth2015omen}. In these scenarios, the adversary is in possession of one or more password hashes obtained from a target system and their primary goal is to obtain the plaintext version corresponding to the respective password/s hash~\cite{blocki2018economics}. In general, a powerful adversary can employ a series of heuristics~\cite{hashcatmask,hashcatrules,hashcatwiki,johntherippermarkov,weir2009password,durmuth2015omen} using a combination of tools~\cite{hashcatweb,johntheripperweb} While doing so, the adversary seeks to avoid a worst-case scenario in which they would need to enumerate all potential guesses, i.e., brute-force.

In this work, we narrow down our experiments and comparisons to \emph{deep generative models} for offline password guessing.\footnote{Our experiments are limited in scale by GPU access. All experiments are done on a single consumer GPU. Scaling is a crucial factor to improve performance and training on larger leaks.} We acknowledge the field is broader than that (e.g., Markov models and online guessing), but we restrict ourselves to comparable architectures. Our goal is to provide additional tools in the password-guessing landscape rather than establishing a default go-to architecture. We summarize our contributions as follows:

\begin{itemize}[topsep=0pt]
    
    \item We introduce PassGPT, an autoregressive transformer that obtains state-of-the-art results in password generation and generalization to unseen datasets. 
    
    \item We show how PassGPT enables a novel approach to password generation under arbitrary constraints: \emph{guided password generation}.
    
    \item We examine password probabilities under PassGPT and how they align with strength. We discuss how this metric could be used to improve current strength estimators.
    \item We present PassVQT, a similar architecture enhanced with vector quantization to increase generation perplexity.
\end{itemize}

%% file: sections/03-background.tex
\section{Background and Related Work}
\label{sec:background}

In this paper, we make heavy use of LLMs and generative AI. In this section, we introduce the concepts relevant to generative models (Section~\ref{sec:dgm}) and \emph{transformer} models (Section~\ref{sec:transformers}). We conclude discussing progress in password guessing and strength estimation, focusing on works that use deep generative models (Section~\ref{sec:dgmpasswords}).

\subsection{Deep Generative Models}
\label{sec:dgm}

Deep generative models are a class of deep learning (DL) techniques designed to \emph{autonomously} grasp the characteristics underlying a set of samples from a distribution, i.e., training set, and to generate new samples from that distribution~\cite{goodfellow2014gans,tomczak2022deep}. The primary distinction between the two major categories of generative models is how they represent probability distributions. Generative models can be either: (1) implicit or (2) explicit. \emph{Implicit} models do not estimate the distribution of the training data directly; rather they learn a function that generates samples similar to the ground truth. The most notable example is Generative Adversarial Networks (GANs)~\cite{goodfellow2014gans}. \emph{Explicit} models, on the other hand, explicitly model the underlying distribution of the training data, that can be later accessed~\cite{sutskever2014sequence}. Our models fall under this second category.

\subsubsection{Generative Adversarial Networks}~\cite{goodfellow2014gans}. 
GANs consist of two main components: (1) a generator $G(\mathbf{z}; \theta_g) : \mathbb{R}^{n} \to \mathbb{R}^{n}$, a neural network that takes in random noise from a prior $p_z$ and generates samples resembling the training data and (2) a discriminator $D(\mathbf{x}; \theta_d) : \mathbb{R}^{n} \to [0, 1] $, also a neural network, trained to distinguish between training samples and outputs from the generator.
\begin{equation}
    \resizebox{0.91\columnwidth}{!}{$\min_G\max_D V(D,G) = \mathbb{E}_{\mathbf{x}\sim p_{\text{data}}(x)}[\log D(\mathbf{x})] + \mathbb{E}_{\mathbf{z}\sim p_{z}(z)}[\log(1-D(G(\mathbf{z})))]$}
    \label{eq:gan}
\end{equation}

Both $G$ and $D$ are trained adversarially in a zero-sum game until the generator produces samples that are indistinguishable from real ones (see Equation~\ref{eq:gan}). GANs can approximate sharp distributions and generate high-quality samples without defining a likelihood function.

\subsubsection{Autoregressive Generative Models}
\label{sec:arms}

 (AGMs). These explicit generative models make it possible to sample from the target distribution. They do so by specifying a probability density function over the data and decomposing it into the product of conditionals via the chain rule of probability. These conditionals can be parametrized using neural networks with parameters $\theta$ that take as input the preceding entries in the sequence (see Equation~\ref{eq:autoregressive}). This explicit definition makes it possible to train these models using \emph{maximum likelihood estimation} unlike implicit generative models. Our models match this definition.

 \begin{equation}
\label{eq:autoregressive}
    p(\mathbf{x}) \approx p(\mathbf{x}; \theta) = \prod_{i=0}^n p(x_i | x_0, \dots x_{i-1}; \theta) = \prod_{i=0}^n p(x_i | x_{<i}; \theta).
\end{equation}

\subsection{Transformers}
\label{sec:transformers}

Choosing the right neural network to model conditional probabilities in deep autoregressive models has received a lot of attention in recent research~\cite{bond2021deep}. Two commonly used architectures are \emph{recurrent neural networks} (RNNs)~\cite{rumelhart1985learning} and \emph{transformers}~\cite{vaswani2017attention}. Transformers are the most successful because of faster, more stable, and parallelizable training~\cite{vaswani2017attention}. 

Transformers rely entirely on the attention mechanism~\cite{sutskever2014sequence} to model dependencies within the input sequence regardless of distance. The original transformer~\cite{vaswani2017attention} consists of an encoder and a decoder, both made up of multiple layers of self-attention and feed-forward neural networks. The main difference between the encoder and the decoder lies in how they consume the input. The encoder uses all information in the input sequence to generate a latent representation for each token, while the decoder can only use information from previous tokens. Recent work has proposed using only the decoder for autoregressive language modeling, where words are generated conditioned only on previous ones. GPT models \cite{radford2018improving,brown2020language} have revolutionized NLP by relying solely on transformer decoders.

\subsection{Related Work}
\label{sec:dgmpasswords}

This section focuses primarily on the use of deep generative models for \emph{password guessing} and \emph{password strength estimation}.

\subsubsection{Password Guessing}
\label{ssec:password_guessing}

 is a widely studied class of attacks~\cite{morris1979password,feldmeier2001unix}, where the adversary either has a limited number of guesses (\emph{online password guessing}) or is already in possession of a copy of the password hashes and needs to break them (\emph{offline password guessing}). In both these scenarios, the adversary seeks to crack passwords before they run out of budget, i.e., the number of tries in an online service, or computing resources available for offline guessing.

 The research community has explored different approaches to guessing passwords efficiently.

Tools like Hashcat~\cite{hashcatweb} or John the Ripper~\cite{johntheripperweb} employ heuristics, such as mangling-rules, dictionary attacks, association attacks, hybrid attacks, and more~\cite{hashcatwiki,hashcatslowcandidates,hashcatrules,hashcatmask,johntherippermarkov}. Further work in the domain has proposed and evaluated the use of Markov models~\cite{narayanan2005fast,durmuth2015omen}, probabilistic context-free grammars (PCFG)~\cite{weir2009password}, (deep) neural networks~\cite{melicher2016fast,ciaramella2006neural,pasquini2021improving,pal2019beyond}, or composition of techniques~\cite{xu2021chunk}. Our work focuses on the use of generative deep neural networks.

\subsubsection{Deep Generative Models for Password Guessing.}
To the best of our knowledge, PassGAN~\cite{hitaj2019passgan} is the first work implementing generative models, in particular GANs, for password guessing. PassGAN uses the improved Wasserstein GAN (IWGAN)~\cite{gulrajani2017improved} to learn the underlying distribution of the RockYou password leak~\cite{rockyousource} and then evaluates the model performance on additional leaks like LinkedIn data~\cite{linkedinwiki}.
Pasquini et al.~\cite{pasquini2021improving} suggested an improved version of PassGAN by adding random noise to the input representation to improve training stability. Follow-up work has explored different architectures obtaining similar results. For instance, PassFlow employs normalizing flows instead of GANs~\cite{pagnotta2022passflow}.

\subsubsection{Password Strength Estimation} aims to define a password robustness metric against guessing~\cite{ur2012does,carnavalet2015large}. Similarly to password guessing, this has resulted in a variety of different approaches such as Markov models~\cite{castelluccia2012adaptive,dell2010password}, PCFGs~\cite{weir2009password}, or neural networks~\cite{melicher2016fast}. Our work uses the \emph{lightweight} estimator \texttt{zxcvbn} \cite{wheeler2016zxcvbn}, as recommended by Carnavalet et al. \cite{carnavalet2015large}. 

%% file: sections/04-experimental-setup.tex
\section{Experimental Setup}
\label{sec:setup}

In Section \ref{sec:dgmpasswords} we introduced the central problem we are exploring: password guessing. This section presents the datasets (Section \ref{sec:datasets}) and novel architectures (Section \ref{sec:models}) used throughout our experiments.

\subsection{Datasets}
\label{sec:datasets}

We chose datasets previously utilized in password guessing work and security research\footnote{This work makes use of publicly available password datasets. We consider this practice to be ethical and consistent with prior security research, e.g., \cite{hitaj2019passgan,pasquini2021improving,melicher2016fast}.} that enable comparison of our techniques. The diverse characteristics of these password sets enhance the robustness of our evaluation and demonstrate generalization capabilities. Table \ref{tab:datasets} summarizes the key information about each dataset. The largest leaks that we consider for training are RockYou and LinkedIn, as done by previous work~\cite{pasquini2021improving,hitaj2019passgan,pagnotta2022passflow}.

We split the previous datasets into training and test sets using the same approach as PassGAN~\cite{hitaj2019passgan} and follow-up work~\cite{pasquini2021improving}. For RockYou, we take the list of all passwords of at most 10 and 16 characters, respectively. In this leak, passwords may appear more than once. We take 80\% of this list as training data. From the remaining 20\%, we keep as test data all passwords that are not contained in the training split, keeping only passwords with low frequency. The most commonly used password in the test set appears only 7 times in the entire leak. The average frequency of test passwords is 1.03. In comparison, the most frequent password in RockYou --\texttt{123456}-- appears 290,731 times, and the average frequency in the entire leak is 2.28. This method allows us to test our model's generation abilities on low-probability passwords that were not seen during training. 

Since the LinkedIn leak does not provide information about password frequency, we take 80\% as training data and the remaining 20\% for evaluation. We ensure that no password appears in both sets. Additionally, we define \emph{cross-evaluation} test sets by removing RockYou training passwords from the LinkedIn evaluation set and vice versa to evaluate generalization to unseen distributions.

Finally, we also consider the MySpace, phpBB, and Hotmail \cite{rockyousource} leaks as evaluation sets. We perform the analogous cross-evaluation procedure to remove RockYou and LinkedIn training data from all of them.

\begin{table}[]
    \centering
    \caption{Main facts about the datasets used in this work. 
    }
    \label{tab:datasets}    
\begin{tabular}{ccc}
\hline
\textbf{Name}                & \textbf{Unique passwords} & \textbf{Year} \\ \hline
LinkedIn \cite{linkedinwiki} & 60,505,270                       & 2012          \\
RockYou \cite{RockYouWiki,rockyousource}   & 14,344,391                       & 2009          \\
phpBB \cite{rockyousource}               & 184,318                      & 2009          \\
MySpace \cite{rockyousource}              & 37,144                       & 2006          \\
Hotmail \cite{rockyousource}             & 8,931                      & Unknown       \\ \hline
\end{tabular}
\end{table}

\subsection{Our Models}
\label{sec:models}
Transformers are a versatile and broad family of deep-learning models as discussed above in Section \ref{sec:transformers}.
For password guessing, we are interested in autoregressive generative models (see Section \ref{sec:arms}). \emph{PassGPT} and \emph{PassVQT} model the probability of a character in a password, given the previous ones: $p(x_i | x_0,\cdots, x_{i-1}; \theta)$. Sampling sequentially from this distribution can generate likely passwords. Our models operate over a vocabulary, $\Sigma$, comprising 256 UTF-8 characters.

Neural networks require a vector representation of tokens as input. We define a \emph{tokenizer} as a function that maps every character $\sigma$ in the vocabulary to an integer,
\begin{equation}
    \text{tokenizer} : \Sigma \mapsto [0, |\Sigma|-1].
\end{equation}
Then, a vector representation is created for each token $\sigma$ using a one-hot encoding of its image under the tokenizer function. This results in a vector of dimension $|\Sigma|$, with all entries equal to zero and a single entry equal to 1 at position $\text{tokenizer}(\sigma)$. 

\subsubsection{PassGPT}
\begin{figure}[h]
    \centering
    \includegraphics[width=.8\textwidth]{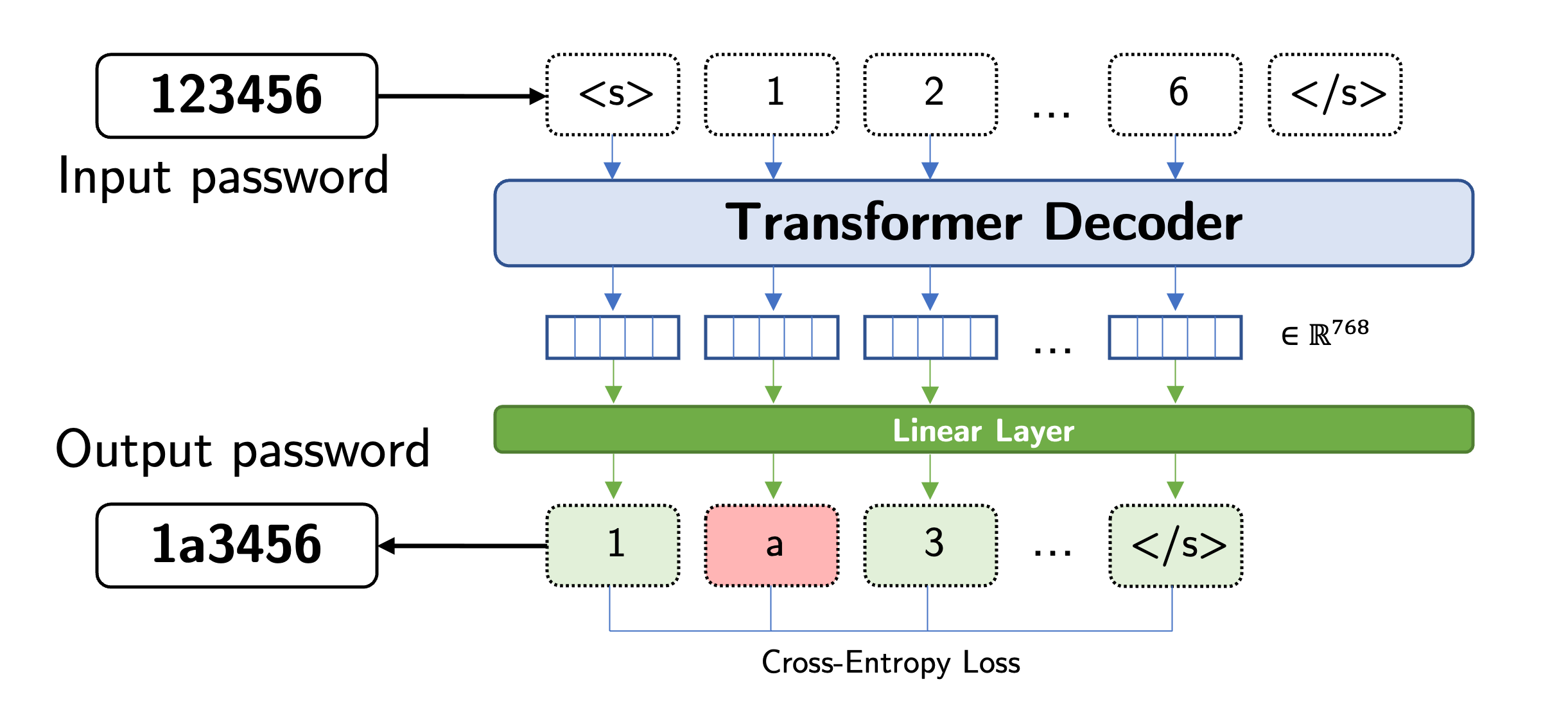}
    \caption{\emph{PassGPT} autoregressively predicts the input character at position $n$ using all previous tokens. Green indicates correct prediction; red indicates incorrect.}
    \label{fig:passgpt}
\end{figure}

, depicted in Figure \ref{fig:passgpt}, is an implementation of the GPT-2 architecture \cite{radford2019language}. GPT models utilize the decoder component of transformers and are trained to predict the next token in a sequence autoregressively. To predict a specific character $x_i$ in a password, the transformer decoder considers only previous characters $x_0, \dots, x_{i-1}$ as input, and outputs a latent vector with dimension $d$ ($d=768$ in our work). This latent vector is then mapped into a real vector of dimension $|\Sigma|$ through a linear layer and further transformed into a probability distribution over the vocabulary using the \emph{softmax} function. The output distribution over the vocabulary represents $p(x_i | x_{<i}; \theta)$. This distribution is optimized using cross-entropy loss with respect to the one-hot-encoding representation of the true character found at that position.

Once the network is trained, it provides us with a parameterized distribution over our vocabulary conditioned on previous tokens, namely, $p(x_i | x_{<i}; \theta)$. For generation purposes, we can start from the start-of-password token, \texttt{<s>}, and find $p(x_1 | x_0 = \texttt{<s>})$. This assigns a probability to every character in our vocabulary to be the first token in the password. If we sample from this distribution, we can fix the first character and repeat the process to find the second one by computing $p(x_2 | x_0, x_1)$. The sampling process for a password finishes when the end-of-password token, \texttt{</s>}, is sampled from the distribution at any given step. Unlike training, this process is sequential.

Our implementation of PassGPT uses the HuggingFace library \cite{wolf-etal-2020-transformers} and has the following specifications: 12 attention heads, 8 decoder layers, and GeLU activation \cite{hendrycks2016gaussian}. Additionally, we train all models for 1 epoch with AdamW optimizer and a starting learning rate of 5e-5 with linear decay during training. 

\subsubsection{PassVQT} 
\begin{figure*}[t]
     \centering
     \begin{subfigure}[c]{0.48\textwidth}
         \centering
         \includegraphics[width=\textwidth]{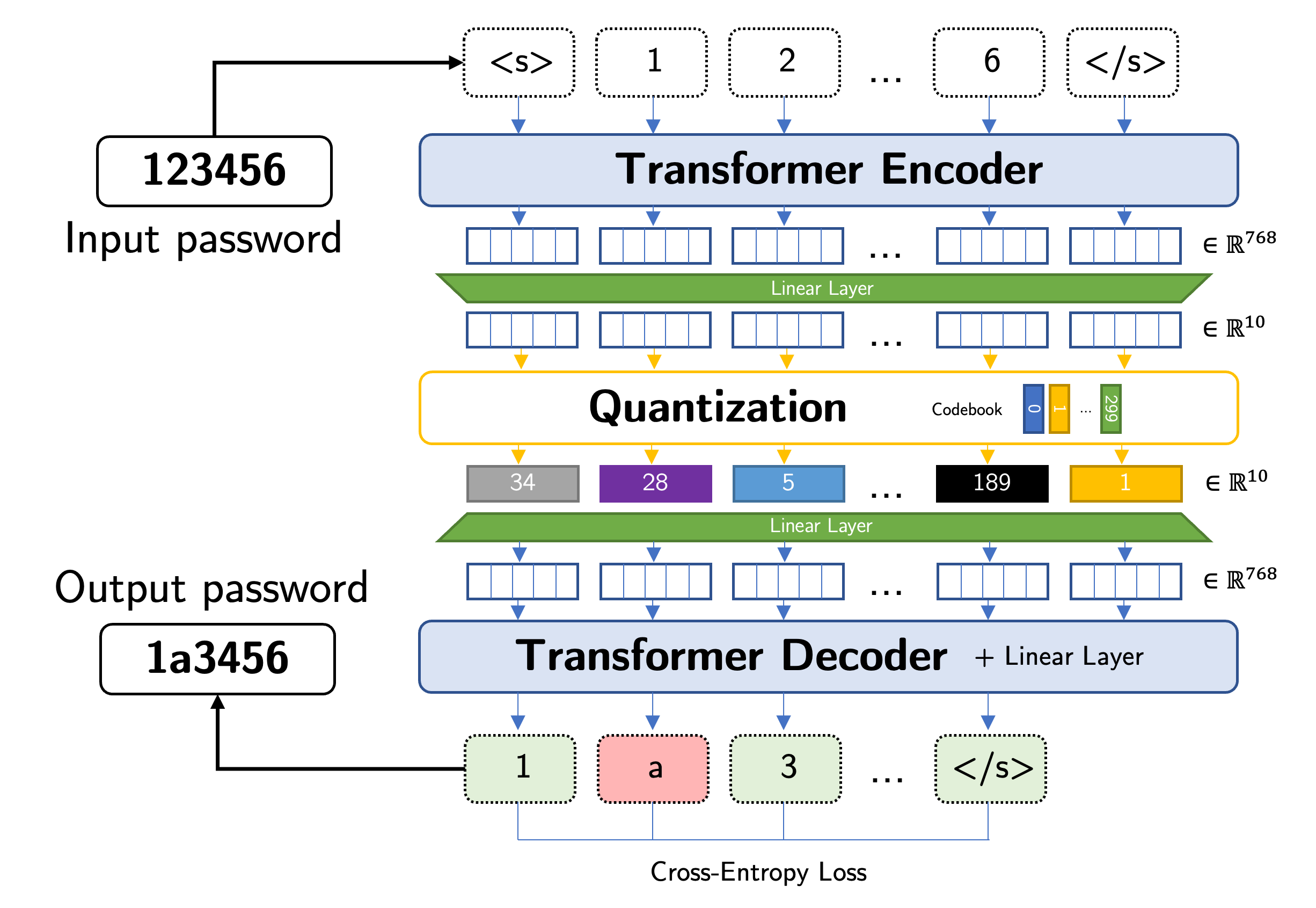}
         \label{fig:vqt1}
     \end{subfigure}
     \hfill
     \begin{subfigure}[c]{0.48\textwidth}
         \centering
         \includegraphics[width=\textwidth]{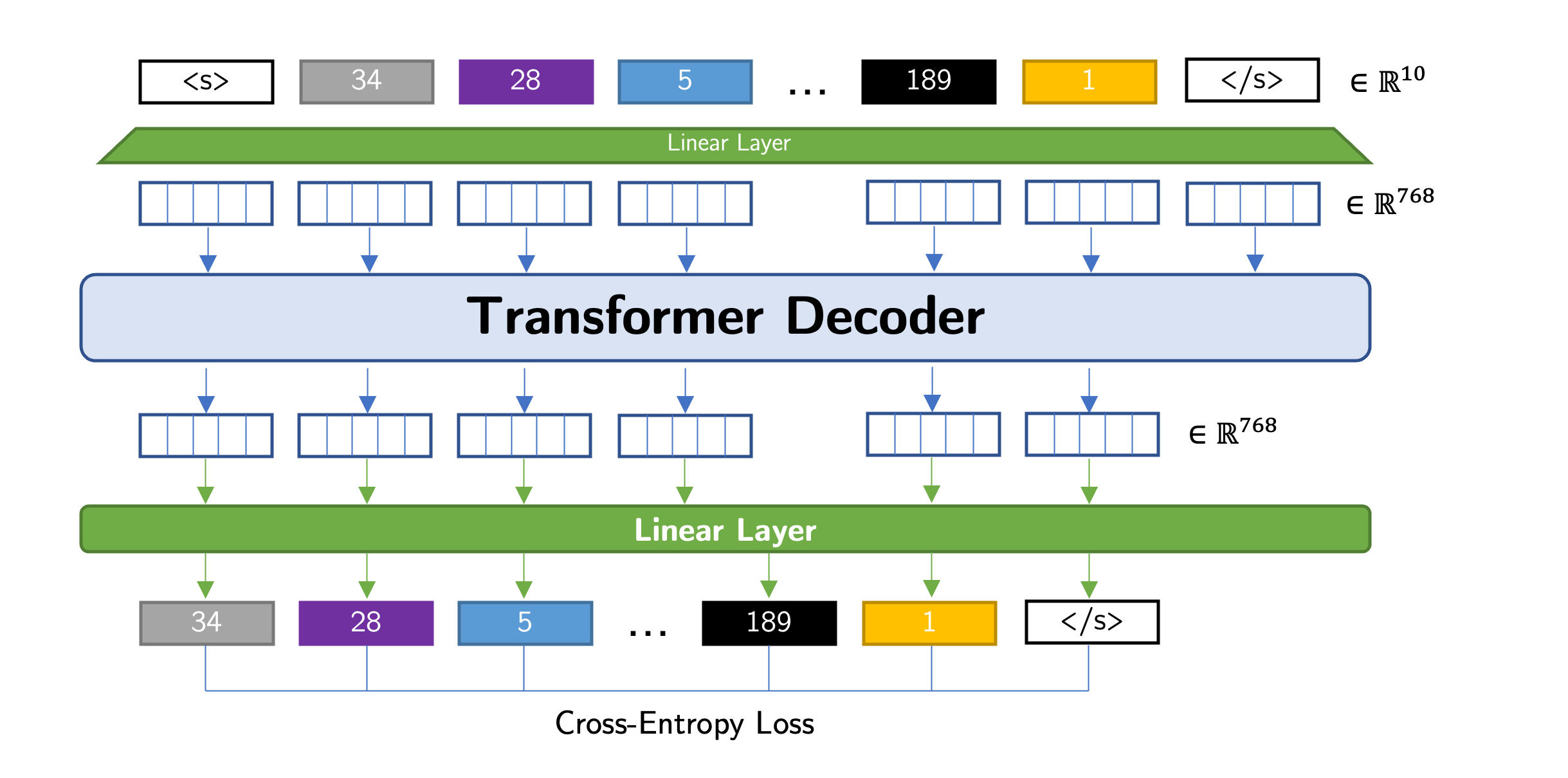}
         \label{fig:vqt2}
     \end{subfigure}
     \caption{Overview of PassVQT showing (left) an end-to-end model trained to compress passwords into a quantized latent space, where each code represents a fixed vector of dimension 768 and (\textbf{right}) an autoregressive GPT model that parameterizes the conditional distribution of indices. The latter is trained once the first one has converged and is required for generation. Transformer decoders in both models are independent.}
     \label{fig:vqt}
\end{figure*}

enhances the transformer architecture with vector-quantization of the latent space. In the computer vision domain, this has been shown to improve sample quality~\cite{yu2021vector}. PassVQT follows the architecture designed by Yu et al.~\cite{yu2021vector}. While modeling the same conditional distribution as PassGPT, we aim to assess whether quantization can provide any additional benefits. In this architecture, depicted in Figure \ref{fig:vqt}, a transformer encoder maps each input token to a latent representation with a dimension of 768. This latent representation is then mapped to 10 dimensions using a linear layer and quantized using k-means and a codebook with $N$ entries. The quantized 10-dimensional vectors are mapped back to 768 dimensions through a linear layer and serve as input to a transformer autoregressive decoder. This decoder is trained to reconstruct the input password character by character, using only the quantized representations for previous tokens.

We carried out a hyperparameter search by minimizing the reconstruction loss on the RockYou leak's training split. Our findings showed that deeper encoder and decoder structures offered better results, with a codebook size of 300 providing the best performance. PassVQT employs a transformer encoder and a GPT-2 decoder with 12 attention heads and 8 layers, respectively. It was implemented on using the HuggingFace library \cite{wolf-etal-2020-transformers} and trained end-to-end with AdamW optimizer and a starting learning rate of 5e-5 with linear decay.

Once the encoder-decoder network has converged, the model can reconstruct input passwords from a compressed quantized latent representation. If we model the distribution of latent codes, we can sample from it to produce a likely sequence of codes, which the decoder can then transform into likely passwords. For this, we train an autoregressive \emph{codes model} over the quantized representation of the training dataset. During inference, we create new passwords by sampling sequences of codes from the \emph{codes model} and transforming them into passwords using the original decoder. The encoder is no longer needed.

%% file: sections/05-evaluation.tex
\section{Evaluation}
\label{sec:evaluation}

Our foremost contribution focuses on password generation. This section compares PassGPT and PassVQT with state-of-the-art deep generative models and demonstrates their generalization to different datasets without the need for further training. We also examine the probabilities and entropies of passwords under PassGPT to provide insights into its capabilities and modeled distribution. Finally, we analyze the alignment of these probabilities with password-strength estimators and discuss how they can be used to improve strength estimation.

\subsection{Password Generation}
\label{sec:generation}

For a fair comparison with PassGAN~\cite{hitaj2019passgan} and its improved version (PassGAN+) \cite{pasquini2021improving}, we train PassGPT and PassVQT using 80\% of passwords of at most 10 characters in the RockYou leak. The evaluation of the generation process is determined by the percentage of passwords from a disjoint test set that the models can generate. In this case, the test set comprises the unique passwords in the remaining 20\% of the RockYou leak that are not in the training set.

\begin{table}[t]
\centering
\caption{Percentage of RockYou test set (10 characters or fewer) guessed from $10^7$ generations. Models are trained on either all passwords or unique entries from RockYou.}
    \label{tab:uniquevsall}
\begin{tabular}{ccc}
\hline
\textbf{Architecture}             & \textbf{Trained on}    & \textbf{\% Test set guessed} \\ \hline
\multirow{2}{*}{PassGPT} & Unique        & \textbf{4.25\%}   \\           
                         & All passwords & 0.53\%               \\\hline
\multirow{2}{*}{PassVQT}     & Unique        & 0.14\%   \\           
                         & All passwords & \textbf{2.86\%}               \\ \hline
\end{tabular}
\end{table}

We consider two variations of the training set: (1) unique passwords and (2) all occurrences. PassGPT demonstrates superior generalization when trained on unique passwords, as detailed in Table \ref{tab:uniquevsall}. Conversely, PassVQT experiences difficulty generating in-distribution passwords when trained on unique entries but significantly improves upon incorporating their absolute occurrences. 

We sample increasingly large pools of password guesses from PassGPT (trained on unique passwords) and PassVQT (trained on all passwords) and calculate the percentage of the RockYou test split they recover. Results in Table \ref{tab:comparisongan} show that PassGPT outperforms all other models. It recovers 41.9\% of the test set among $10^9$ guesses, whereas state-of-the-art GAN models matched 23.33\%. PassVQT performance surpasses that of the original PassGAN and stays close to that of the PassGAN improved version.

\begin{table*}[]
    \centering
    \caption{Percentage of the RockYou test split (<10 characters) matched by samples from various models. PassGAN* stands for the improved PassGAN presented in \cite{pasquini2021improving}. Results for GANs were taken directly from original papers \cite{hitaj2019passgan,pasquini2021improving} and not reproduced.}
    \label{tab:comparisongan}    
\begin{tabular}{ccccc:c}
\hline
\textbf{Guesses} & \textbf{PassGAN} & \textbf{PassGAN}* & \textbf{PassVQT} & \textbf{PassGPT} & \hspace{0.2em}\textbf{PassGPT $\cup$ PassVQT} \hspace{0.2em} \\ \hline
$10^4$           & 0.01\%            & -                  & 0.004\%      & 0.01\%            & 0.01\%    \\                                
$10^5$           & 0.05\%            & -                  & 0.05\%       & 0.05\%            & 0.10\%      \\                              
$10^6$           & 0.38\%            & -                  & 0.45\%       & \textbf{0.50\%}   &                  0.93\%   \\                      
$10^7$           & 2.04\%            & -                  & 2.90\%       & \textbf{ 4.25\%}  &                     6.39\%  \\                    
$10^8$           & 6.73\%            & 9.51\%             & 10.30\%      & \textbf{19.37\%}  &                 22.70\%   \\                  
$10^9$           & 15.09\%           & 23.33\%            & 21.46\%      & \textbf{41.86\%}  &              44.66\%                             \\ \hline
\end{tabular}
\end{table*}

\begin{figure}[]
    \centering
    \includegraphics[width=0.6\textwidth]{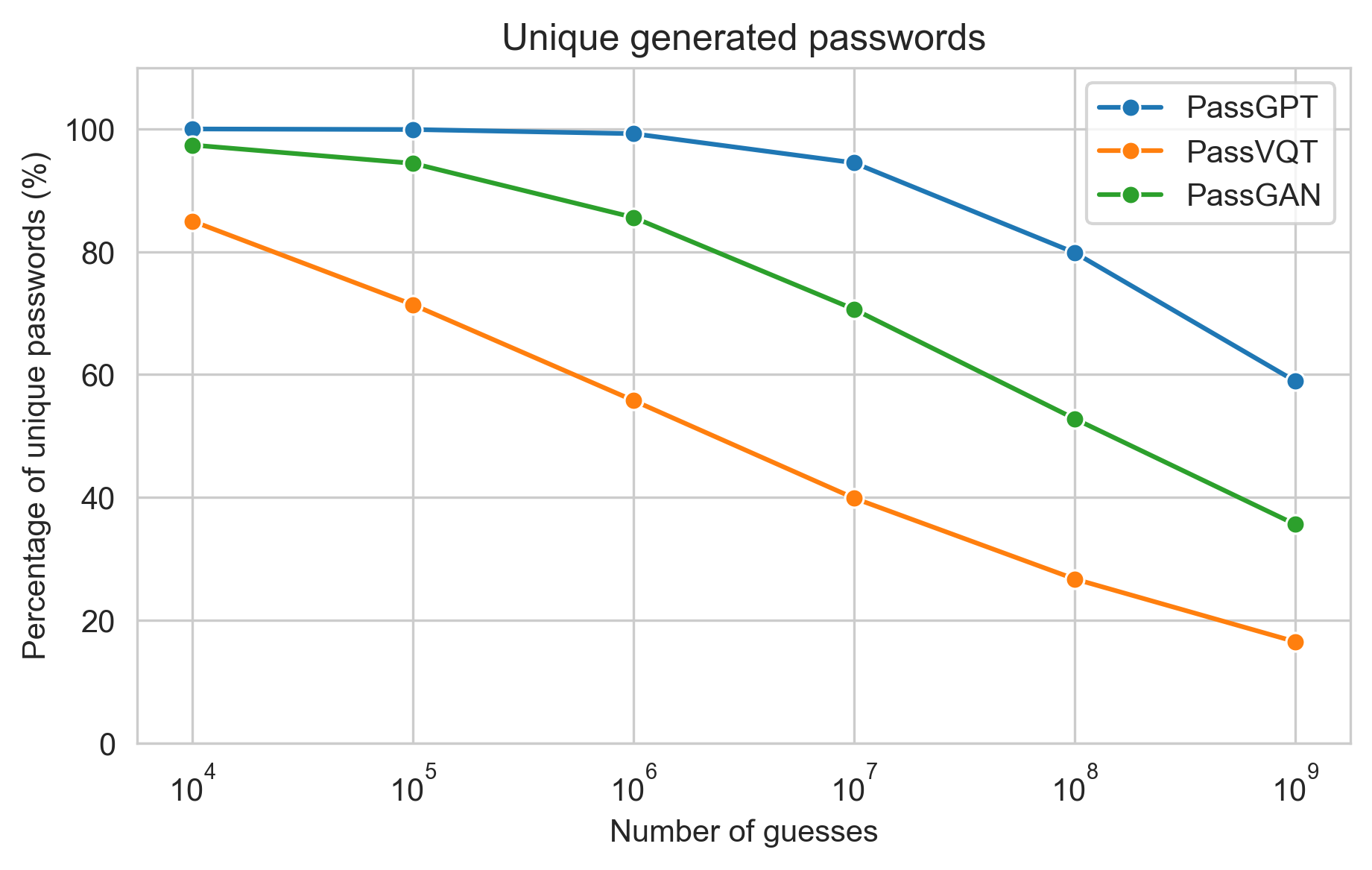}
    \caption{Log-Linear plot of unique passwords generated by different architectures.}
    \label{fig:unique}
\end{figure}

Another important factor in password generation evaluation is the ability to generate novel and distinct samples. We compared the percentage of unique passwords generated by our models to those from PassGAN; results are shown in Figure \ref{fig:unique}. PassGPT retains the highest percentage of unique passwords (60\%), whereas PassVQT drops to 20\% of unique passwords among $10^9$ guesses. Since PassVQT was trained on all occurrences of passwords, under its distribution common passwords are more likely to be generated, reducing the number of novel passwords. PassGAN stays between them with approximately 40\% unique entries.

\subsection{Generalizing to Longer Passwords and Unseen Distributions}

Our models outperform state-of-the-art deep generative models in a common setup. To further evaluate the effectiveness of our models, we extend the modeling to longer passwords, which are more representative of real-world distributions. We train PassGPT and PassVQT on passwords with up to 16 characters (including longer passwords primarily increases entries that are difficult to guess). We again train the models using both unique and all occurrences of the data. PassGPT, as before, performs best when trained on unique samples. Surprisingly, PassVQT now obtains better performance when trained on unique passwords. After training on this new distribution, models retain similar accuracy. From $10^8$ guesses, PassGPT and PassVQT recover $15.5\%$ and $8.57\%$ of the test set, respectively, compared to $19.37\%$ and $10.30\%$ in the 10-character setting. From now on, we will focus on 16-
character models for a richer analysis.

To assess the models' generalization to unseen password distributions, we test them on leaks different from the  RockYou leak. Although users tend to reuse passwords, they are likely to vary based on platform and year of creation\cite{das2014tangled,bailey2014statistics}. We first analyze the LinkedIn leak, which is the largest of our samples and was obtained 3 years after the RockYou leak. To determine how well RockYou models generalize, we benchmark them against a PassGPT model trained solely on 80\% of LinkedIn data. We take the remaining 20\% as the test set after excluding any passwords present in the RockYou training set. This results in a test set of over 11M unique passwords unseen by any of the models during training. We evaluate the models' performance by determining the percentage of test passwords generated by each architecture. RockYou models achieve comparable results to the LinkedIn-trained PassGPT, as shown in Table \ref{tab:linkedingen}, demonstrating the ability of autoregressive models to parameterize rich distributions that generalize beyond the training leak without the need for retraining.

\begin{table}[]
    \centering
    \caption{Percentage of passwords from the LinkedIn test split guessed. Columns indicate training distribution. The test set does not contain passwords in the RockYou training set.}
    \label{tab:linkedingen}    
\begin{tabular}{cccc}
\hline
\multicolumn{1}{l}{\multirow{2}{*}{Guesses}} & \multicolumn{2}{c}{PassGPT} & PassVQT   \\
\multicolumn{1}{l}{}                         & \hspace{2em}\textbf{RockYou}\hspace{1em}  & \hspace{1em}\textbf{LinkedIn}\hspace{1em} & \hspace{1em}\textbf{RockYou}\hspace{2em} \\ \hline
$10^4$                                       & 0.001\%           & 0.001\%            & 0.001\%          \\
$10^5$                                       & 0.012\%           & 0.010\%            & 0.012\%          \\
$10^6$                                       & 0.11\%            & 0.10\%             & 0.13\%           \\
$10^7$                                       & 1.03\%            & 0.94\%             & 1.10\%           \\
$10^8$                                       & 6.03\%            & 6.80\%             & 5.41\%           \\ \hline
\end{tabular}
\end{table}

Finally, we evaluate the RockYou and LinkedIn models to determine which training leak leads to a better generalization. The models are tested on three additional datasets: phpBB, MySpace, and Hotmail (refer to Section \ref{sec:datasets}), after removing passwords present in either the RockYou or LinkedIn training sets. The results are shown in Table \ref{tab:generalization}. RockYou models exhibit superior performance, with password recovery rates of $9.45\%$, $11.39\%$, and $7.22\%$ from $10^8$ guesses.

\begin{table}[]
    \centering
        \caption{Percentage of phpBB, MySpace, and Hotmail leaks generated by PassGPT trained on RockYou, compared with PassGPT trained on LinkedIn. Evaluation is performed on the entire leak after entries contained in the RockYou training set are removed.}
    \label{tab:generalization}    
\begin{tabular}{|c|ccc|ccc|}
\hline
\multirow{2}{*}{\textbf{Guesses}} & \multicolumn{3}{c|}{\hspace{0.5em}PassGPT trained on RockYou\hspace{0.5em} } & \multicolumn{3}{c|}{\hspace{0.5em}PassGPT trained on LinkedIn\hspace{0.5em} }     \\
                                  & \textbf{phpBB}       & \textbf{MySpace}      & \textbf{Hotmail}     & \textbf{phpBB} & \textbf{MySpace} & \textbf{Hotmail} \\ \hline
$10^4$                            & 0.002\%       & 0\%        & 0\%          & 0\%            & 0\%              & 0\%              \\
$10^5$                            & 0.02\%     & 0\%         & 0.02\%       & 0.008\%        & 0\%              & 0\%              \\
$10^6$                            & 0.20\%     & 0.22\%      & 0.18\%   & 0.10\%         & 0.10\%           & 0.05\%           \\
$10^7$                            & 1.80\%     & 2.06\%      & 1.24\%   & 0.77\%         & 0.94\%           & 0.61\%           \\
$10^8$                            & 9.45\%    & 11.39\%    & 7.13\%     & 6.02\%         & 6.57\%           & 4.67\%           \\ \hline
\end{tabular}

\end{table}

\subsection{Guided Generation}

We propose a novel approach to password generation: \emph{guided password generation}. Unlike previous deep generative methods that generate passwords as a whole, PassGPT models each token separately, granting full control over each character. This allows the generation process to meet specific constraints. Some examples of these constraints are: password length, fixed characters (e.g., "a" at first position) and templates (e.g., four lowercase letters and two numbers). This can be achieved by restricting the sampling distribution $p(x_i | x_0, \cdots, x_{i-1})$ to consider only the probability mass assigned to a subset of interest $\Sigma' \subset \Sigma$; for instance, limiting $\Sigma'$ to lowercase letters for the first four tokens. The resulting password generation is guided by these constraints while still being likely under the modeled password distribution. Table \ref{tab:guidedgen} shows various templates and their corresponding generations produced by PassGPT.
\vspace{-1.8em}
\begin{table}[H]
    \centering
    \caption{\emph{Guided generation} examples from PassGPT. Templates formatted using \texttt{l} for lowercase, \texttt{u} for uppercase, \texttt{d} for digit, \texttt{p} for punctuation, and \texttt{*} for any character.}
    \label{tab:guidedgen}    
    \begin{tabular}{cccc} \hline
       \texttt{llllll}  &  \texttt{lllldd} & \texttt{ullppdd} & \texttt{uuuu**dd} \\ \hline
       orange  & manb13 & Nms\_\_12 & PARLA198 \\
       iluvma  & sall89 & Zac\&\&09 & CELAN777 \\
       gikiyd  & lowm12 & Chl@(18 & QWER1234 \\ \hline
    \end{tabular}
\end{table}

\subsection{Probabilities and Entropies Estimates by PassGPT}
\label{sec:probs}

One of the main advantages of autoregressive models is having access to an explicit representation of the modeled distribution. We exploit this property to provide further intuitions behind the PassGPT generation process.\footnote{For PassVQT, this is not possible, as the modeled distribution is in the codebook space, and different codes can lead to the same password generation.} The probability of a password is estimated as the product of the conditional probability for each sampled character, which is more conveniently represented as the log probability (Equation \ref{eq:logprob}). Furthermore, the entropy measures the uncertainty in the model for each token and is calculated according to Equation \ref{eq:entropy}.
\begin{equation}
    \log_{10} p(\mathbf{x}; \theta) = \sum_{i=1}^n \log_{10} p(x_i | x_1, \dots x_{i-1}; \theta).
    \label{eq:logprob}
\end{equation}

\begin{equation}
    H(X_i) =  \sum_{x_i \in \Sigma} p(x_i | x_{<i}; \theta) \cdot \log_{2} p(x_i  | x_{<i}; \theta).
    \label{eq:entropy}
\end{equation}

We computed the log probability and entropy for every position in all unique passwords (<16 characters) in the RockYou dataset using PassGPT. Examples of passwords with different probabilities under the model can be found in Appendix \ref{sec:appprob}. Figure \ref{fig:entropy_PassGPT} depicts the distribution of the entropy for characters found at specific positions in passwords of length 16. The entropy of the first character is slightly above five; it is constant since $p(x_1 | x_0 = \texttt{<s>})$ is equal across passwords. The median entropy decreases as we move towards the last positions because the model reduces uncertainty as more characters are observed. Figure \ref{fig:logprob_PassGPT} illustrates how the log-probability of passwords decreases with length, with the median log-probability dropping by approximately 1 unit for each additional character. This corresponds to an average probability of 0.1 for each new character.

We can analyze password probabilities under the model compared to brute-force search. Our vocabulary $\Sigma$ contains 256 characters. Therefore, the log-probability of discovering a password of length 3 through brute force can be approximated as $\log_{10}(1 / 256^3) \approx -7.5$. This value is close to the median log-probability under PassGPT. However, the utility of generative methods becomes evident when we deal with longer passwords that are computationally infeasible to uncover through exhaustive search. For instance, the log-probability of successfully recovering a 16-character password using brute-force attacks is $-38.5$. In contrast, the median log-probability under PassGPT hovers around $-18$. This indicates that finding a 16 character password using PassGPT is approximately $10^{20}$ times more likely than relying on random guessing.

\begin{figure}[t]
     \centering
     \begin{subfigure}[b]{\textwidth}
         \centering
         \includegraphics[width=\textwidth]{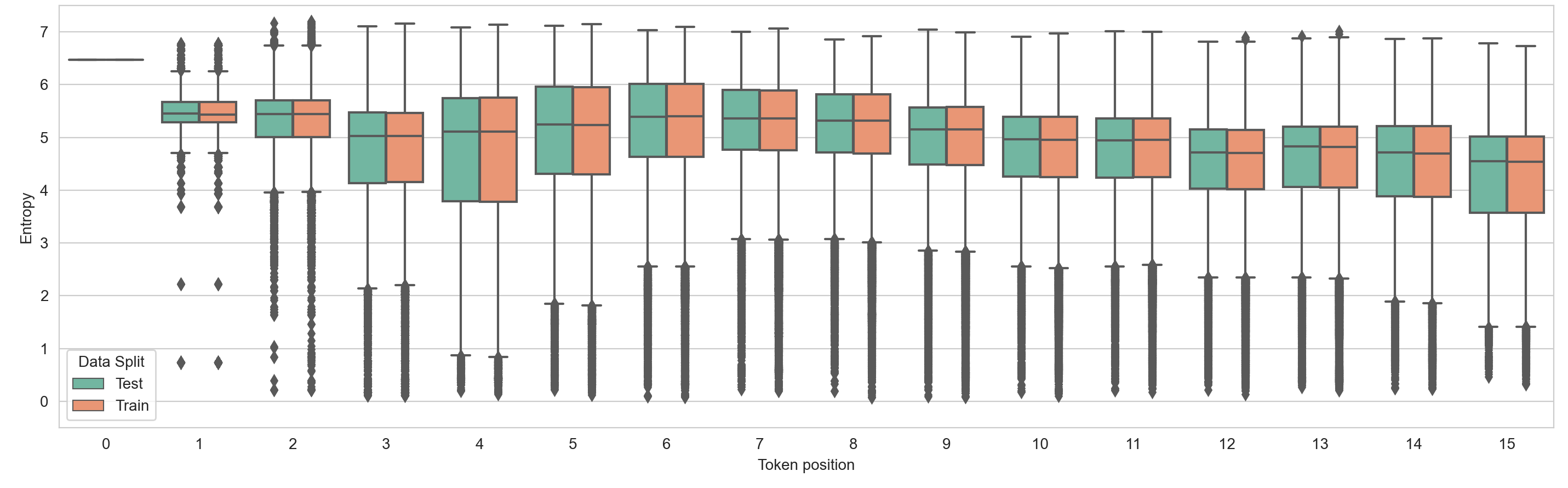}
         \caption{Entropy distribution per token for passwords of length 16 under PassGPT.}
         \label{fig:entropy_PassGPT}
     \end{subfigure}
     \hfill
     \vspace{0.4em}
     \begin{subfigure}[b]{\textwidth}
         \centering
         \includegraphics[width=\textwidth]{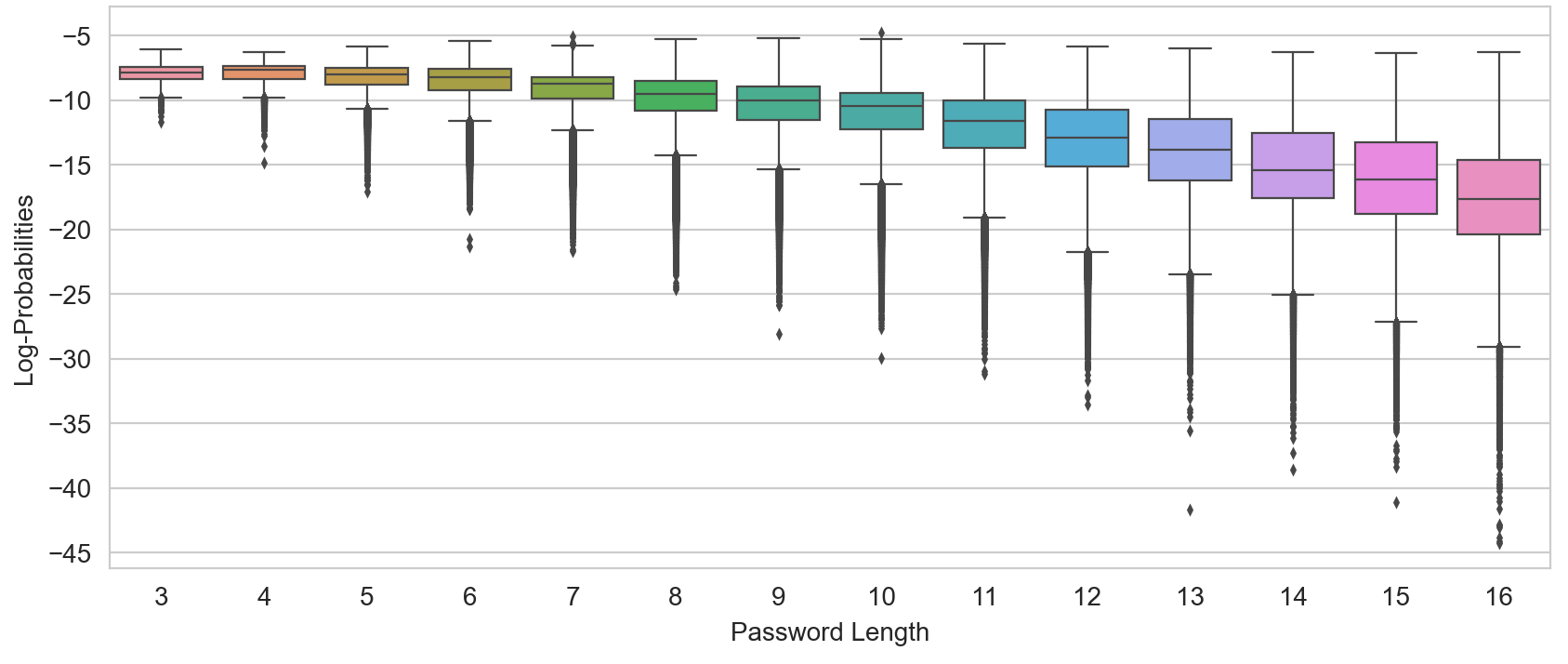}
         \caption{Log-probabilities with respect to length for all passwords in RockYou.}
         \label{fig:logprob_PassGPT}
     \end{subfigure}
        \caption{Entropy and log-probability of passwords in the RockYou leak under PassGPT}
\end{figure}

\subsection{Discussion: PassGPT vs PassVQT}

We wrap up this section with a brief discussion about the main differences between PassGPT and PassVQT, and when to use each of them. Details can be found in Appendix \ref{sec:appcomp}. Overall, these models can surpass state-of-the-art deep generative models and generalize to unseen distributions. Focusing on models trained on 16 characters, we can highlight several differences:

\begin{enumerate}[topsep=0pt]
    \item PassVQT generates longer passwords than PassGPT.
    \item PassGPT guesses weaker passwords, while PassVQT matches stronger passwords.
    \item PassGPT generates more unique passwords than PassVQT: 84\% vs 76\%. 
    \item PassGPT can generate passwords faster than PassVQT: 12h vs 24h to generate $10^8$ samples on 1 NVIDIA RTX3090.
\end{enumerate}

All things considered, PassGPT seems better at modeling the actual leaked distribution and generating in-distribution samples. On the other hand, PassVQT is "more imaginative" and creates stronger passwords with a similar distribution to that of the leaked file. Nevertheless, the sampling process of both models can be tweaked to pursue specific goals. For instance, if we want to reduce out-of-distribution samples, we can perform top-k sampling for each character, considering only the most likely tokens under the model to avoid long-tail passwords. Similarly, to incentivize the generation of stronger and less likely passwords, we can increase the temperature of the \emph{softmax} function, or avoid sampling from the top-k most likely tokens.



\begin{figure}[t]
    \centering
    \includegraphics[width=\textwidth]{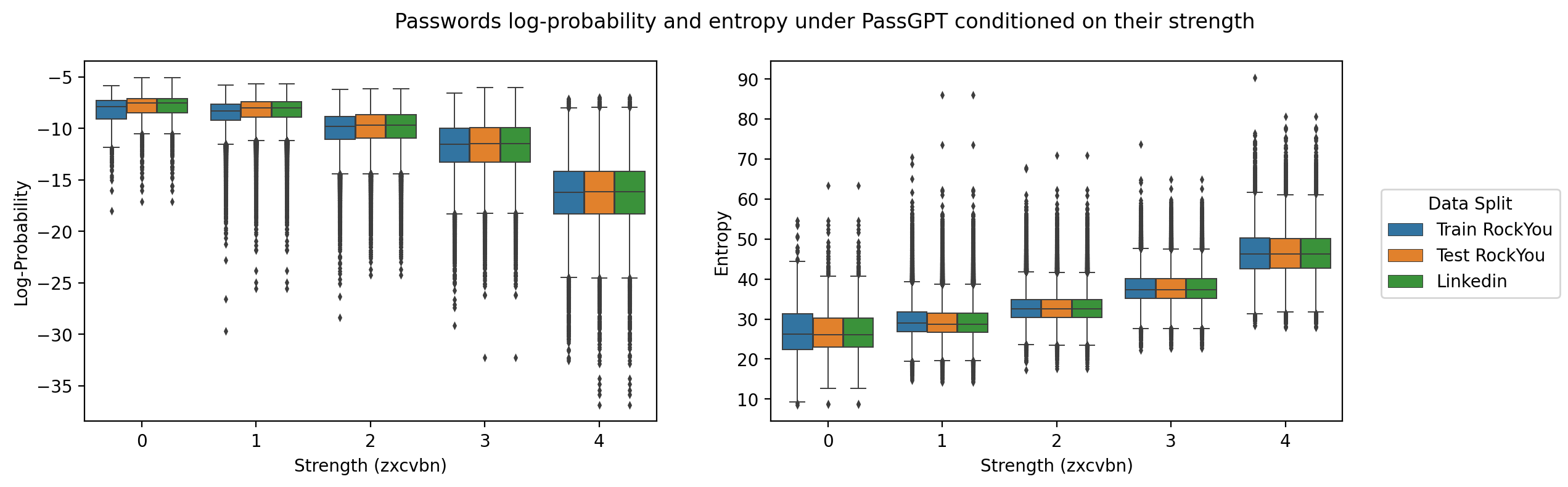}
\caption{Log-probability and entropy under the model according to password strength (\texttt{zxcvbn})}
\label{fig:strength}
\end{figure}

\subsection{Password Strength Estimation}
\label{ssec:pwd_strength_estimation}

In the previous section, we carried out a comprehensive analysis of the fundamental characteristics of the probabilities and entropies of passwords in PassGPT. In this section, we delve deeper into the relationship between probability, entropy, and password strength to gain a better understanding of the modeled distribution. For each unique password in the RockYou and LinkedIn leaks, we calculate its log-probability and entropy under PassGPT and its strength as determined by \emph{zxcvbn} \cite{wheeler2016zxcvbn}. This method assigns a score ranging from 0 (very weak) to 4 (very strong). The distribution of log-probabilities and entropies for each strength score is illustrated in Figure \ref{fig:strength}. The results show that PassGPT assigns lower probability and higher entropy to stronger passwords, thereby demonstrating that weak passwords are more likely in the modeled distribution.

Finally, we conduct a manual examination of outliers in the distributions to better understand when PassGPT does not align with \emph{zxcvbn}. Our analysis revealed three distinct phenomena where the model assigns low probabilities to passwords that are considered weak by \emph{zxcvbn}. These examples are hard to model for PassGPT, but easy to detect using dictionary attacks.

\begin{enumerate}[topsep=0pt]
    \item Pattern repetition. These passwords are composed of a sequence that is repeated several times. Examples: ":X:X:X:X:X:X", "qwertqwertqwert".
    \item Replacement of characters in common words by similar symbols. A dictionary attack is successful to find these passwords. Examples: "k1m83rly" (from "kimberly") or "r00sevelt" (from "roosevelt").
    \item Reversed words. These can also be easily detected by \emph{zxcvbn} but are unlikely under the model distribution. Example: "llabtooF" (from "Football").
\end{enumerate}

 On the other hand, there are very strong passwords, according to \emph{zxcvbn}, that obtain high probabilities under the model. We can also identify predominant phenomena:
\begin{enumerate}[topsep=0pt]
    \item Passwords containing non-English words. \emph{zxcvbn} tries to decompose them as English words unsuccessfully. For example, the password "teamomiamorcito" is formed by the Spanish words "te amo mi amorcito" ("I love you my love"). However, \emph{zxcvbn} parses it as "team", "omi", "amorcito". 
    \item Love-related passwords. "iloveyou" is one of the most common passwords in RockYou. When analyzing passwords with strength 4 that obtained high probabilities, we found copious variations of it. Examples: "ilovematt4eva", "ilovetoby4eva", "ilovetyler4ever", "ilovehotmail", "iloveyousomuch". The suffixes "4ever" and "4eva" are very common among these passwords.
\end{enumerate}

It is crucial for strength estimators to minimize the number of false negatives, i.e., classifying passwords that can be guessed by any existing technique as strong. Our analysis revealed instances of very strong passwords with high probabilities under PassGPT, indicating that they are likely to be discovered by such a model. We believe that incorporating the log-likelihood from generative models into existing password strength estimators could provide valuable supplementary information and improve the accuracy of these systems for high-stake scenarios.

%% file: sections/07-conclusions.tex
\section{Conclusions}
\label{sec:conclusions}

In this work, we investigated the use of large language models to model password distributions without explicit supervision. We introduced two autoregressive architectures that model the conditional distribution of characters based on previous ones: PassGPT and PassVQT. PassGPT might be preferable because it provides access to an explicit probability distribution, is simpler, and provides faster generation. However, PassVQT might still be helpful for scenarios where we want to express more variability and generate more complicated passwords that are still close to the training distribution.

Advantages of autoregressive models over state-of-the-art GAN generators include \emph{guided password generation} and access to an explicit probability distribution. We have analyzed how the log-probabilities of passwords under PassGPT align with their strength, and how this metric could be used to mitigate limitations in strength estimators.

Overall, this work seeds many promising research directions in the field of password modeling using large language models that are to be explored by future research.

%% file: sections/10-appendix.tex
\section{Passwords at the quantiles of the distribution}
\label{sec:appprob}

\begin{figure}[H]
    \centering
    \includegraphics[width=\textwidth]{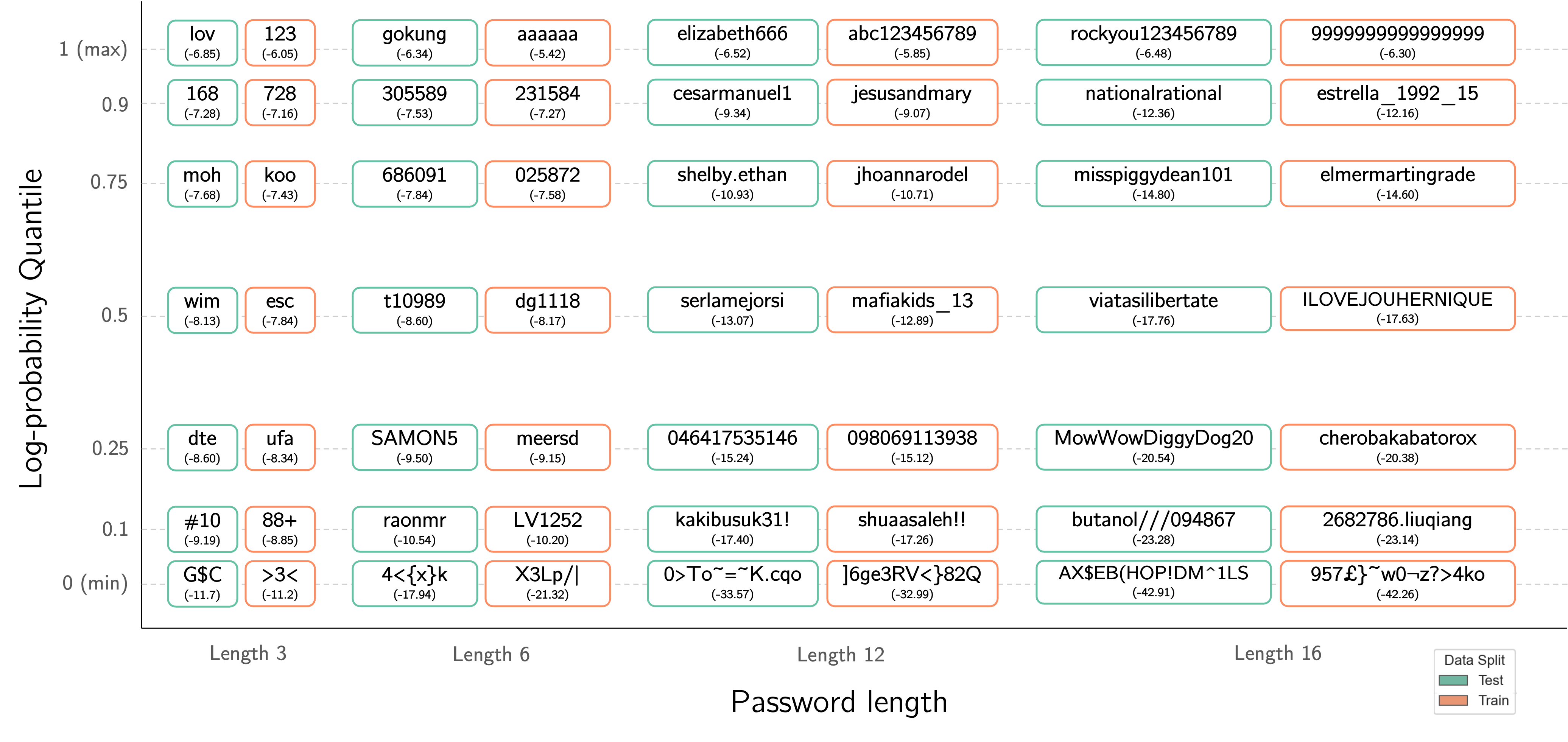}
    \caption{Passwords in RockYou located at the quantiles of the probabilities distribution for different password lengths. The lower the quantiles, the more unlikely the password is under the model.}
    \label{fig:logprobquantiles}
\end{figure}

In our analysis in Section~\ref{sec:probs}, we examined password probabilities under PassGPT in relation to password strength. PassGPT assigns lower probabilities to stronger passwords and higher probabilities to weaker ones. In Figure~\ref{fig:logprobquantiles}, we illustrate some passwords along with their log-probabilities under PassGPT for visual exploration.





\section{PassGPT vs PassVQT}
\label{sec:appcomp}

\begin{figure}[H]
    \centering
    \includegraphics[width=0.6\textwidth]{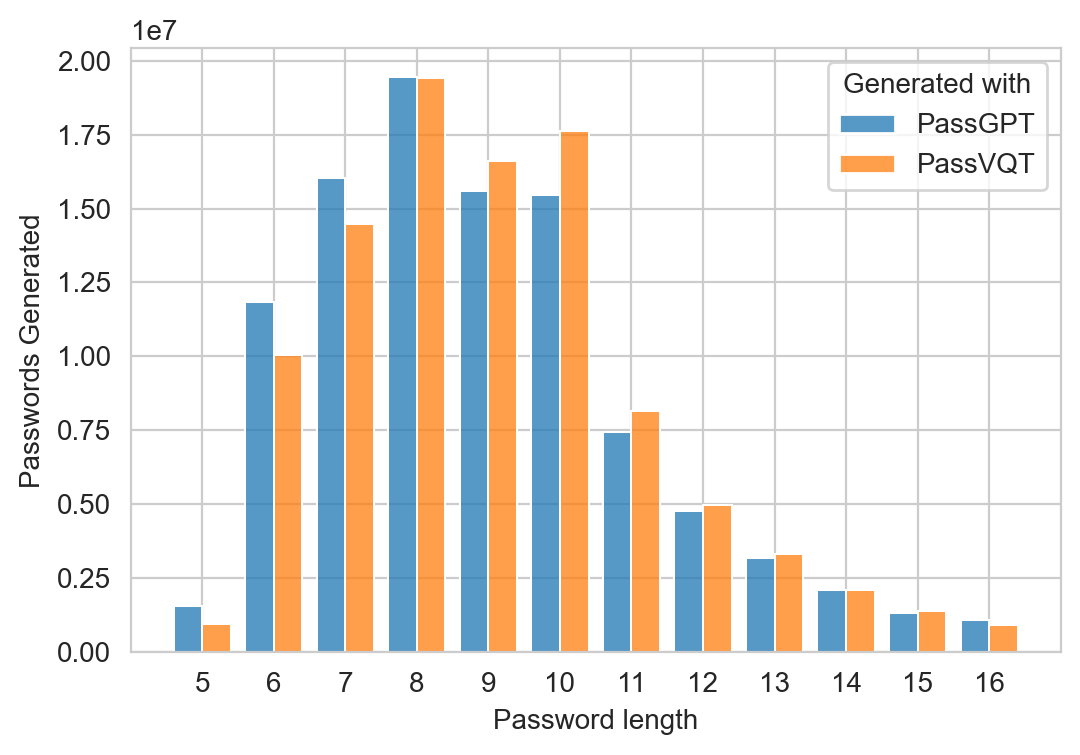}
    \caption{Histogram of generated password length by each model on a subset of $10^8$ samples.}
    \label{fig:genlength}
\end{figure}

This section includes a detailed comparison between PassGPT and PassVQT generations. Table \ref{tab:strengthgen} illustrates how many passwords are guessed by each method conditioned on their strength. Finally, Figure
\ref{fig:genlength} depicts a histogram of the length of passwords generated by each architecture. PassVQT overall tends to generate longer and more difficult passwords, but PassGPT better fits the distribution of easy passwords, improving its overall performance.

\begin{table}[H]
\centering
\caption{Detailed statistics on the matched passwords by different models from RockYou test set (<16 characters) with respect to their strength. Passwords are only considered once.}
\label{tab:strengthgen}
\begin{adjustbox}{max width=\textwidth}
\begin{tabular}{|c|c|ccc|c|}
\hline
\multirow{2}{*}{\textbf{Strength}} & \multirow{2}{*}{\textbf{Total}} & \multicolumn{3}{c|}{\textbf{Guessed by}}                & \multirow{2}{*}{\textbf{Not guessed}} \\ \cline{3-5}
                                   &                                 & PassGPT $\cap$ PassVQT & PassGPT          & PassVQT        &                                       \\ \hline
0                                  & 2,035                           & 41 (2.0\%)          & 481 (23.7\%)     & 28 (1.4\%)     & 1485 (73\%)                           \\
1                                  & 752,137                         & 150,945 (20.1\%)    & 111,704 (14.9\%) & 66,004 (8.8\%) & 423,484 (56.3\%)                      \\
2                                  & 926,826                         & 45,337 (4.9\%)      & 50,704 (5.5\%)   & 53,782 (5.8\%) & 777,003 (83.8\%)                      \\
3                                  & 558,029                         & 4,501 (0.8\%)         & 7,831 (1.4\%)    & 8,832 (1.6\%)  & 536,865 (96.2\%)                      \\
4                                  & 158,635                         & 88 (0.06\%)         & 194 (0.1\%)      & 152 (0.09\%)   & 158,201 (99.7\%)                      \\ \hline
 & 2,397,662 & 200,912 (8.38\%) & 170,914 (7.12\%) & 128,798 (5.37\%) & 1,897,038 (79.12\%)\\ \hline

\end{tabular}
\end{adjustbox}

\end{table}

